\RequirePackage{fix-cm}
\documentclass[twocolumn,natbib]{svjour3}          % twocolumn
\smartqed  % flush right qed marks, e.g. at end of proof
\usepackage{graphicx}
\usepackage{amsmath}
\usepackage{amsfonts}
\usepackage{booktabs}       % professional-quality tables
\usepackage{xcolor}
\usepackage{colortbl} 
\usepackage{algorithm2e}
\RestyleAlgo{ruled}
\usepackage{multirow} % table multirow
\usepackage{float} % figure position

\def\ie{{\em i.e.}}
\def\eg{{\em e.g.}}
\def\etal{{\em et al.}}

\usepackage{balance}

\usepackage[colorlinks, citecolor=blue]{hyperref}

\makeatletter
\def\cl@chapter{\@elt {theorem}}
\makeatother
\usepackage{cleveref}
\Crefname{algocf}{alg.}{algs.}
\Crefname{algocf}{Algorithm}{Algorithms}

\newcommand{\rom}[1]
{\MakeUppercase{\romannumeral #1}}

\usepackage[htt]{hyphenat}

\begin{document}
	
	\title{Adapting Vision-Language Models from Iconic to Inclusive for Multi-Label Recognition Without Labels}
	% \subtitle{Do you have a subtitle?\\ If so, write it here}
	
	
	\author{
		Cheng Chen$^{\dagger}$ \and
		Jingyu Zhou$^{\dagger}$ \and
		Yifan Zhao$^{*}$ \and
		Jia Li$^{*}$
	}

	
	\institute{
		$^{\dagger}$ These authors contributed equally to this work. \\
		$^{*}$ Correspondence should be addressed to Y. Zhao and J. Li. \\
		Cheng Chen \at
		State Key Laboratory of Virtual Reality Technology and Systems, SCSE \& QRI, Beihang University, Beijing 100191, China. \\
		\email{chencheng1@buaa.edu.cn}
		\and
		Jingyu Zhou \at
		State Key Laboratory of Virtual Reality Technology and Systems, SCSE \& QRI, Beihang University, Beijing 100191, China. \\
		\email{JingyuZhou2004@buaa.edu.cn}
		\and
		Yifan Zhao \at
		State Key Laboratory of Virtual Reality Technology and Systems, SCSE \& QRI, Beihang University, Beijing 100191, China.\\
		\email{zhaoyf@buaa.edu.cn}
		\and
		Jia Li \at
		State Key Laboratory of Virtual Reality Technology and Systems, SCSE \& QRI, Beihang University, Beijing 100191, China.\\
		\email{jiali@buaa.edu.cn}
	}

	\maketitle
	
	\begin{abstract}
		Understanding multi-label images remains a challenging task in computer vision. With the rapid progress of vision-language multimodal learning, vision-language models (VLMs) enable zero-shot recognition without labeled data. However, due to their intrinsic design, these models often prioritize the most iconic object and omit other contextual positives. This intrinsic bias conflicts with the nature of multi-label learning, thereby limiting their applicability. In this work, we propose an unsupervised framework that adapts VLMs from iconic recognition toward inclusive understanding, enabling label-free multi-label image recognition. Our approach consists of two key stages, ``cutting'' and ``sewing'': In the cutting stage, we present the multi-sampling response estimator to prevent the model from concentrating only on one single object. In the second sewing stage, the multi-object blend adaptation is introduced to adjust the labels to better conform to the multi-label distribution while preserving the intrinsic characteristics of the original model within only one epoch. Extensive experiments show that our framework significantly outperforms existing unsupervised approaches on four public datasets, even surpassing several representative weakly supervised baselines. These results demonstrate the potential of adapting pre-trained VLMs for more comprehensive visual understanding without manual annotations. Our code is publicly available at \url{https://github.com/iCVTEAM/TailorCLIP}.
	\end{abstract}
	
	\section{Introduction}\label{sec:introduction}
	Multi-label image recognition aims at predicting multiple visual objects within a single image, which constitutes a fundamental prerequisite for numerous downstream tasks and practical applications, including object detection \citep{Pathiraja2023detection,Wu2023openDetection,Ma2023openDetection,Liu2023denseDetection}, segmentation \citep{Xu2023sideadapter, Liang2023openclipseg, Zhao2023semiSegmentation, Liu2023zeroshotSegmentation} and retrieval \citep{Xie2023CLIPRetrieval,Lee2023structuralRetrieval,Saito2023Pic2WordRetrieval,Sain2023CLIPSketchRetrieval}. 
	Significant progress in multi-label learning is achieved by the construction of large-scale domain-specific datasets and corresponding semantic annotations. However, annotating all candidates for multi-label images (\eg, over 80 categories for one instance) is extremely challenging not only for its difficulty but also for substantial labor consumption. To alleviate such challenges, previous works \citep{Kim2023gapOfPartialLabel,Zhang2023learnLongTailedPartial,Xia2023holisticMultiLabel,Chen2023MLXray,Rajeswar2022multiLabelAmbiguity,Ben-Baruch2022PartialMLSelectiveLoss,Kim2022largeLossML,chen2023semantic,Chen2022SST,Pu2022SARB} try using partially annotated datasets instead of fully annotated ones, and make steady improvements in closing the performance gap between fully supervised and weakly supervised learning.
	
	\begin{figure}[htbp]
		\centering
		\includegraphics[width=.8\linewidth]{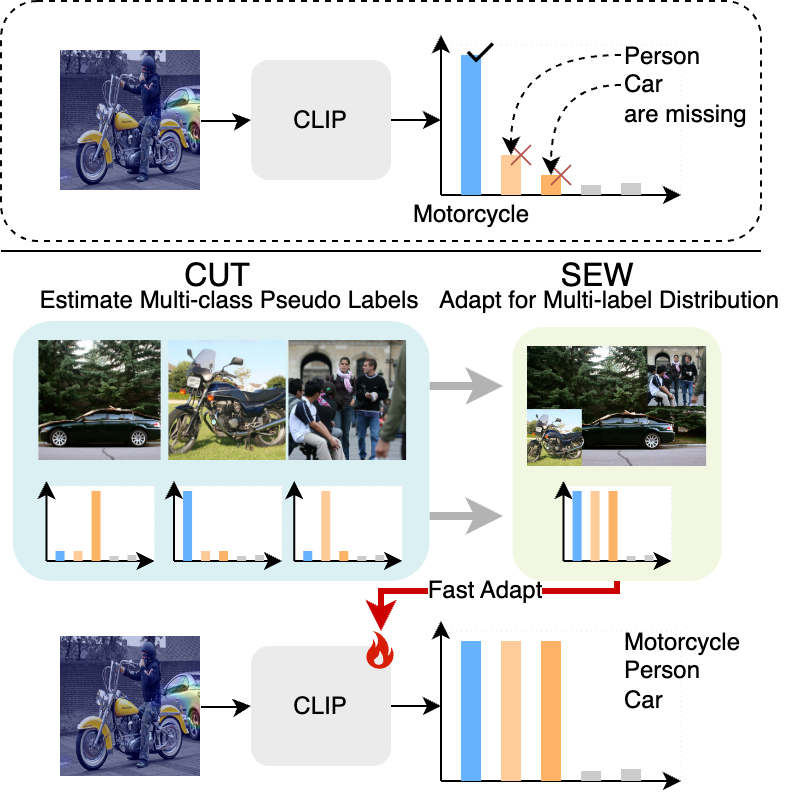}
		\caption{Illustrations of our motivation. Vision-language models are naturally trained for one-positive prediction, unsuitable for unsupervised multi-label learning. Our unsupervised framework has two stages. i) \textbf{Cutting}: we cut and discover local object dictionaries to prevent the model from focusing only on one object. ii) \textbf{Sewing}: we ``sew'' these local dictionaries into a new dataset better conforming to multi-label distributions while preserving intrinsic characteristics.}
		\label{fig:header}
	\end{figure}
	
	With the booming of vision-language multimodal learning, vision-language models (VLMs) have increasingly exhibited promising potential in recognizing unseen objects, a capability also known as zero-shot learning. For example, to recognize unseen images, one can simply use natural language prompts like \textit{`a photo of class'} with pretrained VLMs to classify images without any training process on the downstream data. This capability shows superior generalization on other subtasks, such as semantic segmentation \citep{Xu2023sideadapter,Liang2023openclipseg}, captioning \citep{Zeng2023ConZICCLIPCaptioning,Ramos2023SmallCapCLIPCaptioning} and retrieval \citep{Sain2023CLIPSketchRetrieval, Xie2023CLIPRetrieval, Saito2023Pic2WordRetrieval}. Though these models perform well on classifying single-positive labeled datasets, \eg, ImageNet, in which each image with one salient object is associated with only one label, recent research indicates that VLMs perform poorly on multi-label datasets \citep{Lin2023CLIP-ES, Abdelfattah2023CDUL}. This prediction bias in vision-language models is primarily attributed to their intrinsic design and caption-based contrastive learning, which naturally \textbf{concentrates on the most prominent objects while ignoring the rest inconspicuous ones} that are nonetheless crucial for multi-label learning.
	
	For multi-label recognition, CDUL \citep{Abdelfattah2023CDUL} first uses CLIP to predict and refine outputs by fusing local/global features, and then optimizes another model with pseudo labels in the next phase. However, the potential ability of CLIP models is significantly neglected: 1) it only notices the single-positive limitation of CLIP, without uncovering its intrinsic behavior that CLIP tends to focus on the predominant object while interpreting the rest as context, 2) it relies solely on CLIP logits without further adaptation for multi-label learning, and 3) much CLIP knowledge is lost during distillation.
	
	Following CDUL, several recent works attempt to further exploit multi-label signals. BAC-GCN \citep{jo2025bac} leverages BLIP-2 \citep{li2023blip} captions and a GCN \citep{kipf2016semi} to model class-background relationships. CCD \citep{kim2025classifier} uses CAM to select precise local views and applies a debiasing strategy for CLIP pseudo-labels. TagCLIP \citep{lin2024tagclip} proposes a local-to-global open-vocabulary framework to improve multi-label classification. While these methods partially alleviate the concentration and label limitations of CLIP, they still depend on pseudo-label refinement or auxiliary modules and do not fully exploit the intrinsic multi-label knowledge in pre-trained vision-language models.
	
	To excavate this implicit knowledge, we propose an unsupervised framework adapting vision-language models from iconic recognition toward inclusive understanding, enabling \textbf{label-free multi-label image recognition}. Our approach ``cuts'' and ``sews'' image-level responses from VLMs, evolving them into a rapid multi-label learner on diverse datasets \textbf{without any labels}. As shown in \Cref{fig:header}, the framework contains two stages: \textbf{(i) Cutting}, with a multi-sampling response estimator on local patches generating initial local responses and \textbf{local dictionaries} that elevate non-salient objects; and \textbf{(ii) Sewing}, with a multi-object blend adaptation that adjusts labels to better match multi-label distributions while preserving intrinsic model characteristics, fusing predictions with order-persistent confidence correction. We adopt an EM algorithm to iteratively optimize cutting and sewing with lightweight estimators and adapters, progressively enhancing semantic understanding. The final multi-label model is \textbf{standalone for inference} without relying on the original VLM.

	To sum up, this paper makes the following contributions:
	\begin{enumerate}
		\item We make an experimental analysis to reveal the intrinsic nature of vision-language models (VLMs), \eg, CLIP, on multi-label learning. Besides uncovering their inadequacy, we propose harnessing their zero-shot capability to develop a novel unsupervised multi-label learning framework.
		\item We propose a multi-sampling response estimator to prevent the model from concentrating only on one single object and present multi-object blend adaptation with an order-persistent confidence correction to discover the contextual labels for multi-label training.
		\item We propose an expectation-maximization optimization framework to iteratively evolve the cutting and sewing ability and extensive experiments show that our framework significantly outperforms existing unsupervised approaches on four public datasets, even surpassing several representative weakly supervised baselines.
	\end{enumerate}
	
	The remainder of this paper is organized as follows: \Cref{sect:related} provides the literature review. \Cref{sect:method} gives an analysis of the behavior of vision-language models and describes our proposed framework. The experimental comparisons and detailed ablations are reported in \Cref{sect:expsum}. The conclusions and limitations are summarized in \Cref{sect:conclusion}.

	\section{Related Works} \label{sect:related}
	\subsection{Multi-label Recognition with Incomplete Labels}
	
	Recognizing multiple objects in images is a fundamental task in computer vision region and has been widely investigated \citep{Guo2023textprompt,Zhu2023multiLabelSceneLearning,Zhu2023sceneGraphMultiLabel,Li2023CTMultiLabel,Zhang2022MLContrastive,Liu2022contextualDebias,Zhao2021TDRG,Ridnik_2021_ICCV,Lanchantin2021CTrans}. One crucial challenge in this task is the difficulty and labor consumption of collecting high-quality multi-label data for model training \citep{Cole2021singlePositive, chen2023semantic}, which used to be manageable in multi-class learning. To alleviate this, many methods that train with partial or noisy labels have been proposed \citep{Kim2023gapOfPartialLabel,Zhang2023learnLongTailedPartial,Xia2023holisticMultiLabel,Chen2023MLXray,Rajeswar2022multiLabelAmbiguity,Ben-Baruch2022PartialMLSelectiveLoss,Kim2022largeLossML,Chen2022SST,Pu2022SARB,chen2023semantic}. Kim \etal \citep{Kim2023gapOfPartialLabel} analyze and fix the gaps between models trained on fully labeled and partially labeled data. Xia \etal \citep{Xia2023holisticMultiLabel} leverage memory effects and propose holistic metrics to determine clear labels from noisy ones. SST \citep{Chen2022SST} learns semantic correlations to enhance partial label training, while SARB \citep{Pu2022SARB} explores representation blending to regularize models for better performance.
	
	These methods generally assume that a subset of the training data is fully or partially annotated, which is ``observed,'' while the rest is ``unobserved,'' relying on assumptions such as uniform random sampling of ground truth labels to support regularization or label purification. Building on this line of research, recent approaches further explore alternative strategies: MambaML \citep{zhu2025mambaml} investigates the multi-label state space explicitly to guide recognition; MLC \citep{ma2025correlative} leverages prompt tuning to adapt pre-trained models for multi-label prediction; other works exploit discriminative object representations \citep{zhao2025towards} or open-vocabulary frameworks \citep{tan2025recover} to reduce reliance on dense annotations and improve generalization.
	
	In contrast, our method eliminates the need for any annotations, proposing a fully unsupervised framework that adapts pre-trained vision-language models for multi-label recognition, thereby exploiting their intrinsic multi-label understanding without human labels.
	\vspace{-1em}
	\subsection{Vision-Language Models} 
	Vision-language models build a bridge between visual and textual information, bringing the rich structure of natural language into the visual tasks. Representative methods including CLIP \citep{Radford2021CLIP} performs great success in many downstream tasks, \eg, semantic segmentation \citep{Liu2023zeroshotSegmentation, Xu2023sideadapter, Liang2023openclipseg}, image captioning \citep{Zeng2023ConZICCLIPCaptioning, Ramos2023SmallCapCLIPCaptioning} and retrieval \citep{Sain2023CLIPSketchRetrieval, Xie2023CLIPRetrieval, Saito2023Pic2WordRetrieval}. The methods leveraging CLIP usually do not fine-tune the whole model but adopt prompting and adapters to avoid feature collapse because of few downstream data. CoOp \citep{Zhou2022CoOp}, DualCoOp \citep{sun2022DualCoOp} optimize learnable prompts for each category with labeled data. CLIP-Adapter \citep{gao2021clipadapter} trains a tiny adapter attached to visual or textual encoders with labels. Side Adapter \citep{Xu2023sideadapter} proposes a side network attached to each layer to reuse frozen CLIP features. Liang \etal~\citep{Liang2023openclipseg} fine-tune CLIP on masked images with text description to achieve open-vocabulary segmentation. CLIP-ES \citep{Lin2023CLIP-ES} refines CAM from CLIP to achieve weakly supervised segmentation. These methods \textbf{do not} focus on CLIP zero-shot abilities and require full or partial annotations to fine-tune CLIP for their multi-label tasks, indicating they do not tackle the one-positive limitations of CLIP and avoid the problems with human-annotated data.
	
	\subsection{Training without Labels}
	Benefiting from the generalization capability of the vision-language model CLIP, recognition \textit{without labels} and many other settings such as segmentation, detection are proposed \citep{Liu2023zeroshotSegmentation, Zeng2023ConZICCLIPCaptioning, Sain2023CLIPSketchRetrieval, Saito2023Pic2WordRetrieval, Liu2023channelMultiLabelZeroShot}. ZegCLIP \citep{Zhou2023ZegCLIP} proposes concise modifications to CLIP and brings its prediction to pixel level for segmentation. Li \etal \citep{chen2023visualcue} use visual cues from LLMs to enhance zero-shot relation detection. These methods use prompted CLIP to generate the middle results and refine them with proposed techniques to achieve better performance on various tasks. However, research \citep{Lin2023CLIP-ES, Abdelfattah2023CDUL,kim2025classifier,lin2024tagclip} observes that CLIP does not perform well on multi-label tasks and we propose new framework to excavate the multi-label hidden attributes that are naturally learned by pretrained CLIP.
	
	\subsection{Discussions and Relations}
	For multi-label recognition, CDUL \citep{Abdelfattah2023CDUL} first uses CLIP to predict and refine outputs by fusing local/global features, and then optimize the other model with pseudo labels in the next phase. However, the potential ability of CLIP models is significantly neglected: 1) the previous method only notices the single-positive limitation of CLIP, without further uncovering its underlying mechanism,~\ie, CLIP always focuses on the predominant object and understands the rest as context. These characteristics stem from the image-text paired contrastive pretraining. 2) This method mainly relies on the basic CLIP model as a label predictor without adaptation or further modification on the task of multi-label learning. 3) This method only utilizes the logits of training samples, which results in the loss of a significant amount of CLIP knowledge during distillation. Moreover, BAC-GCN \citep{jo2025bac} uses BLIP-2 \citep{li2023blip} for stronger pseudo labels, relying on a different paradigm, while our method focuses on CLIP’s intrinsic multi-label capacity. To solve these deficiencies, we start with an experimental analysis of CLIP behaviors on multi-label data, find the intrinsic reasons, and propose our unsupervised framework to adapt CLIP into a multi-label learner.

	\section{Approach}\label{sect:method}
	\subsection{Analyses and Overview} \label{sect:methodanalyses}
	
	\subsubsection{Problem Formulation}
	Denote $\mathcal{X}=\{(\mathbf x_i,\mathbf y_i)\}_i^N$ as the training data in fully supervised learning, where $N$ is the data length. For $(\mathbf x_i,\mathbf y_i) \in \mathcal X$, $\mathbf x_i$ is the $i$th image and $\mathbf y_i=[y_{i0},\dots,y_{iC}] \in \mathcal Y$ is the associated full labels to the image $\mathbf x$, where $C$ is the size of candidate labels, $y_{ij}=1$ or $0$ indicates that the objects of $j$th category exist in $\mathbf x$ or not, respectively. The objective of fully supervised learning is to find a function $\mathcal F:\mathcal X' \mapsto \mathcal Y$ for predicting full labels of given images that minimize the risks:
	\begin{equation}
		\min_{\Theta}\mathbb E_{(\mathbf x,\mathbf y) \sim \mathcal X}\xi(\mathcal F(\mathbf x;\Theta),\mathbf y),\end{equation}
	where $\xi$ is the optimization criterion, \eg, binary cross entropy, and $\Theta$ are the learnable parameters. However, \textbf{in unsupervised learning}, $\mathbf y_i$ is \textbf{completely unobserved}, indicating only $\mathcal X'=\{\mathbf x_i\}_{i=1}^N$ is accessible to optimize $\hat{\mathcal F}$:
	\begin{equation}\label{eq:optimization-target}
		\min_\Theta\mathbb E_{\mathbf x \sim \mathcal X'}\xi(\hat{\mathcal F}(\mathbf x;\Theta),\mathcal R(\mathbf x)).
	\end{equation}
	Here we rely on zero-shot CLIP as discovering functions $\mathcal R(\cdot)$, while building pseudo label is an intuitive baseline.
	
	\begin{figure}[t]
		\centering
		\includegraphics[width=\linewidth]{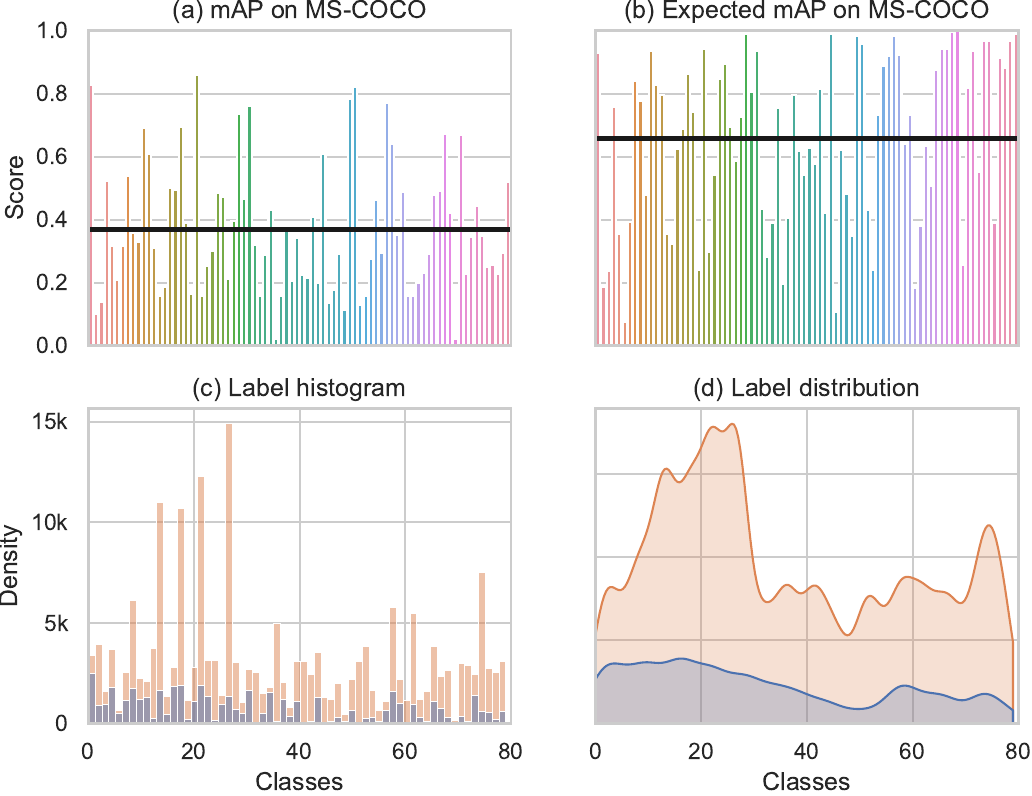}
		\caption{The behavior of vision-language model CLIP on the representative multi-label MS-COCO dataset. (a) and (b) show zero-shot mAP and expected mAP on MS-COCO. The black lines mark average performance. (c) shows the distributions of positive labels of 80 classes from {\color{blue}model prediction} (blue) and {\color{orange}ground truth} (orange). (d) is an estimated continuous curve of (c) for better interpretation.}
		\label{fig:clip-zero-shot-behavior}
	\end{figure}
	\subsubsection{Analysis of One-Positive Vision-Language Models}
	VLMs are typically trained on noisy image-caption data collected from the web by comparing the distances between images and captions in a unified feature space. For an image $\mathbf x$ and multiple texts $\mathcal T=\{t_1,t_2,\dots,t_i,\dots\}$, a typical VLM $\Phi(\mathbf x,\mathcal T)$ predicts the closest one of the texts in space as positive with visual encoder $f_v$ and textual encoder $f_t$, leaving the rest negative:
	\begin{equation}
		\Phi(\mathbf x,\mathcal T)=\mathrm{Softmax}\left([f_t(t_1)~f_t(t_2)~\cdots]^\top
		f_v(\mathbf x)\right).
	\end{equation}
	This design scheme, as well as its accompanying training methods and model structures, \ie, the attention mechanism in both ResNet and ViT backbone, limit VLMs' multi-label reasoning capabilities. In \Cref{fig:clip-zero-shot-behavior} we study zero-shot vision-language models on the MS-COCO dataset and observe that: 1) The model shows inferior performance for multi-label inference. The mAP of the whole 80 classes is $36.9\%$, while the mAP of a naive fine-tuned ResNet-101 achieves $78.5\%$. 2) The expected mAP is $65.8\%$ according to top-1 accuracy if the model performs normally on multi-label recognition tasks, which is much higher than its overall mAP $36.9\%$. 3) \Cref{fig:clip-zero-shot-behavior}(c) and (d) further prove that the model exhibit a significant intrinsic bias in recognition, which is not associated with data.
	%Investigating the outputs, CLIP’s top 3 favorites are ``person'', ``airplane'', ``bus'', and the bottom 3 are ``mouse'', ``toaster'', and ``snowboard''. We draw to our conclusions:
	After investigating the outputs, we find vision-language models are mixed blessings for the multi-label learning tasks:
	\begin{itemize}
		\item Vision-language models show an intrinsic one-positive bias in multi-label tasks, mainly focusing on the most iconic object and harming performance.
		\item This bias ensures the prediction on local image views is plausible, which serves as an iterative guidance during learning. 
	\end{itemize}
	To capitalize on this and circumvent these limitations, we next present our framework, preventing the bias by ``cutting'' and exploiting the bias by ``sewing'' to quickly adapt vision-language models to multi-label tasks without labels.

	\begin{figure*}[t]
		\centering
		\includegraphics[width=0.95\linewidth]{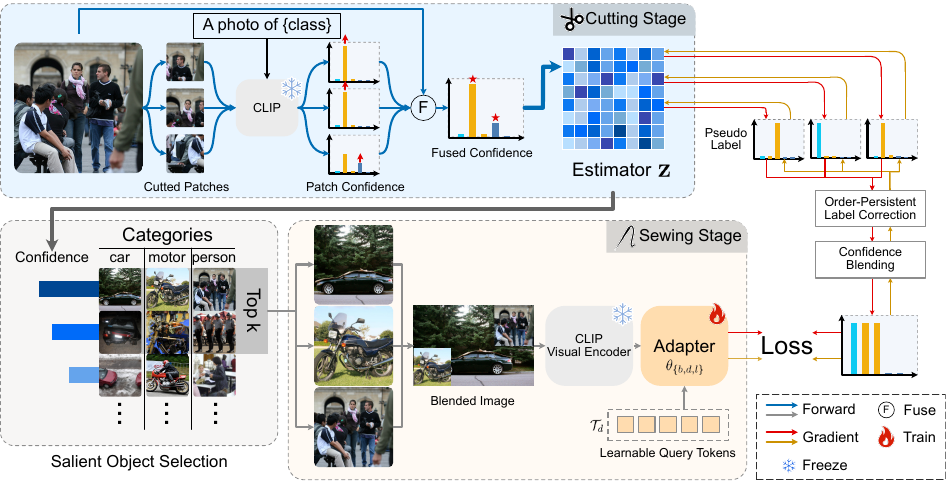}
		\caption{Pipeline of the proposed unsupervised framework. Our approach includes the cutting stage and sewing stage. (\rom{1}) Cutting Stage: we present the multi-sampling response estimator to prevent the model from concentrating only on one single object. The images are sampled and fed to the vision-language model to provide multiple responses. All responses are fused to update the estimator. (\rom{2}) Sewing Stage: according to the initial confidences exported from the estimator, we introduce the multi-object blend adaptation to adjust the labels to better conform to the multi-label distribution while preserving the intrinsic characteristics of the original model. After that, the salient objects are stitched together to generate multi-label images. The plausible pseudo labels are generated by blending their confidence from the estimator. (\rom{3}) Interactive Training. With the estimator and models, the framework simultaneously optimizes the model and corrects confidence errors.}
		\label{fig:framework}
	\end{figure*}
	
	\subsubsection{Framework Overview}
	Our framework includes two key stages: the cutting stage and the sewing stage.
	\begin{enumerate}
		\item In the cutting stage, a multi-sampling response estimator is proposed to mitigate the limitations of vision-language models, prevent them from responding to only salient/iconic objects, and generate pseu\-do labels for unobserved training sets.
		\item In the sewing stage, we introduce multi-object blend adaptation and order-persistent confidence correction. The confidence correction adjusts confidence from the previous stage to regularize labels while respecting its relative orders for knowledge distillation purposes. Multi-object blend adaptation wisely merges responses and labels, adjusting the model outputs to better conform to multi-label data distribution while preserving the intrinsic characteristics of the original model.
	\end{enumerate}
	With the framework in \Cref{fig:framework}, we finally train a lightweight standalone model on fully unlabeled datasets, clearly demonstrating the potential of adapting pre-trained VLMs for more comprehensive visual understanding without any annotations.

	\subsection{Cutting: Multi-sampling Response Estimation}
	
	From the aforementioned discussion, vision-language models are typically designed for one-positive prediction. Such intrinsic characteristics put them at a disadvantage in multi-label learning.
	To mitigate that, we present the multi-sampling response estimator in the first stage to ``cut'' images to prevent the model from concentrating only on one single object and fuse the limited responses for accurate predictions.
	
	Toward this, we first let $\mathbf x_i \in \mathcal X$ be the input for the vision-language model, to get the initial confidence $\mathbf z_i^0$
	\begin{equation}\label{eq:init-z}
		\mathbf z_i^0=\mathcal \mathcal S^{-1}(\Phi(\mathbf x_i,\mathcal T))=[z_{i1}^0,z_{i2}^0,\dots,z_{iC}^0]^\top,
	\end{equation}
	where $\mathcal S^{-1}$ is the inverse function of sigmoid $\mathcal S$. By \Cref{eq:init-z}, the estimator refers to the model's confidence and transfers it to the domain of the sigmoid function, which is more suitable for this task. Denoting $\mathbf Z^k$ the sequences of $\mathbf z^k_i$, the $\mathbf Z^0$ is taken as the initial pseudo labels. However, $\mathbf Z^0$ highly responds to the salient objects in images while ignoring the rest. To solve that without annotations provided, we sample each image $\mathbf x$ multiple times with the ``cutting'' function $\mathcal C(\cdot)$. $\mathcal C(\mathbf x;\rho)$ randomly crops $\mathbf x$ and retains $\rho \in (0,1)$ of the area. The process finally creates
	\begin{align}
		\hat{\mathcal X}'_k=\{\mathcal C(\mathbf x;\rho) \mid \mathbf x \in \mathcal X'\}.
	\end{align}
	\begin{figure}[htbp]
		\centering
		\includegraphics[width=0.96\linewidth]{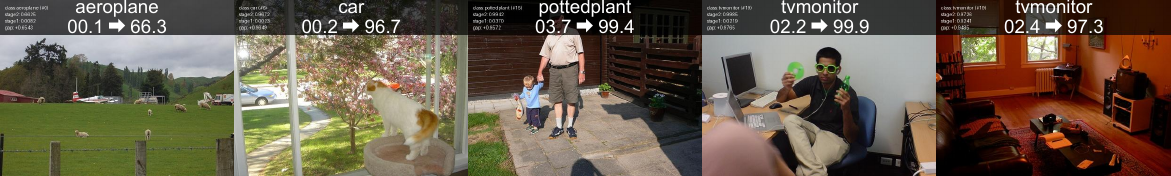}
		\caption{Correction of model prediction bias by the Cutting stage. Before correction, the model is almost unable to predict non-salient objects; after correction, these objects are successfully recovered and the predictions are corrected.}
		\label{fig:stage2}
	\end{figure}
	
	The estimator then uses the vision-language model to predict $\hat{\mathcal X}_k$ and get $\mathbf Z^k$. By sampling, the salient objects and their context are corrupted to prevent the model from focusing on them, and different regions of images $\mathbf x$ are cut out. 
	{Notably, the cutting stage introduces a multi-scale observation effect. Random crops zoom into local regions, allowing the model to capture visual evidence less visible in the global view. Objects with low confidence, such as small, occluded, or fine-grained ones, occupy a larger proportion of the cropped patches, increasing their feature saliency and response scores. This recovers visual cues suppressed by dominant objects and provides complementary evidence for multi-label prediction, as shown in \Cref{fig:stage2}.}
	Considering the similarity measuring mechanism in vision-language models, the regions are separately predicted to provide independent confidences $\mathbf Z^k$. The estimator fuses $\{\mathbf Z^k\}_k$ to get confidence $\mathbf z_i$ for image $\mathbf x_i$ by
	\begin{align}
		\mathbf z_i&=[z_{i1},z_{i2},\dots,z_{iC}]^\top, \\
		\label{eq:7}
		z_{ij}& \gets \begin{cases}
			z_{ij}^0, & \mathrm{initialize} \\
			\max \{z_{ij}, z^k_{ij}-\alpha\}, & k=1,2,\dots
		\end{cases}
	\end{align}
	
	where $\mathbf z_i$ first annotates salient objects and gradually reflects the relative confidences to other objects, hyperparameter $\alpha$ prevents noise from affecting the confidence. Through the estimator, we successfully obtain independent responses of the vision-language model to a multi-label image.

	\subsection{Sewing: Multi-object Blend Adapting}
	
	After the first cutting stage, we get confidence $\mathbf Z$ of unlabeled training data. However, the pseudo labels generated from them are of low quality, and we still have not obtained a standalone model for multi-label recognition. In the sewing stage, we leverage the confidence from the vision-language model, adjusting it to better conform to multi-label distribution to distill the knowledge into a lightweight recognition model with the help of proposed correction and adaptation strategies.
	
	\subsubsection{Order-Persistent Label Correction}
	
	In semi-supervised learning, the confidence is used to generate pseudo labels by thresholding \citep{sohn2020fixmatch,Zhang2021Flexmatch}. The representative method FixMatch in semi-supervised \textit{multi-class learning} leverages the thresholding method to select the most confident label of each sample as positive label $\mathbf z'_{ij}=1$ with the help of data consistency, which can be written as
	\begin{align}
		z'_{ij}=[j={\arg\max}_k z_{ik} \land z_{ik} \geq \tau],
	\end{align}
	where $z'_{ij}$ is the pseudo label of $j$th class of $i$th sample, $\tau$ is the threshold. However, this method is not suitable for multi-label learning, as it is not able to generate multiple labels for one sample.
	
	We argue that the \textit{multi-label} confidence of unlabeled datasets from vision-language models contains fragile intrinsic knowledge, that could be easily corrupted by binarization or thresholding. In a network with sigmoid as the activation function in the last layer and binary cross-entropy as loss, which is commonly used in multi-label learning, the relative confidence order of samples of one class determines the behavior of the network to that class through backpropagation (verified in \Cref{sect:label-correction-ablation}). To generate pseudo labels while leveraging most knowledge for distillation, we propose order-persistent label correction.
	
	Let $\mathbf Z=\{\mathbf z_1,\mathbf z_2,\dots,\mathbf z_N\}$ be the confidence of training data from the first stage. We aim to correct the confidence to generate pseudo labels $\mathbf Z'=\{\mathbf z'_1,\mathbf z'_2,\dots,\mathbf z'_N\}$, where $\mathbf z'_i$ is the pseudo label of $\mathbf z_i$. We propose a correction function $g(\cdot)$ to adjust the confidence $\mathbf z_i$ to $\mathbf z'_i$ by 
	\begin{equation}\label{eq:g}
		g(z_{ij})=\begin{cases}
			\mathcal S^{-1}(\sqrt{\mathcal S(z_{ij})}), & z_{ij}~\text{in top-}\beta~\text{of}~\mathbf z_i \\
			z_{ij}, & \text{otherwise}
		\end{cases}
		% \mathcal S^{-1}(\frac{10\sqrt{\mathbf 100\mathcal S(\mathbf z_{ij})}}{100}), & \text{j in top-k}
		% =\mathcal S^{-1}(\sqrt{\mathcal S(\mathbf z)})
	\end{equation}
	where $\beta$ is a hyperparameter, $\mathcal S$ and $\mathcal S^{-1}$ are sigmoid function and inverse function of it. {When a single image contains more potential objects, a larger $\beta$ helps the model better handle multi-object co-occurrence.}
	
	This ensures that the confidence of existing objects is highlighted in pseudo labels while retaining the consistency of the relative order within the image and the same categories. With $g(\cdot)$, the confidence $\mathbf Z=\{\mathbf z_1,\mathbf z_2,.\\.., \mathbf z_N\}$ are adjusted as pseudo labels $\mathbf Z'=\{\mathbf z'_1,\mathbf z'_2,\dots,\mathbf z'_N\}$ while retaining the knowledge for distillation.
	
	\subsubsection{Multi-object Blend Adaptation}
	With pseudo labels retrieved, we introduce the multi-object blend adaptation to adjust the labels to better conform to the multi-label distribution while preserving the intrinsic characteristics of the original vision-language model.
	
	We explicitly build new multi-label data based on the original training data $\mathcal X'$ and known plausible one-positive confidence $\mathbf Z^{\cdot}$ by ``sewing'' them. This is motivated by the observation that vision-language models are more accurate in recognizing salient objects, as discussed in \Cref{sect:method} and \Cref{sect:cam}. Therefore, we use the initial confidence $\mathbf Z^0$ to construct the patch dataset $\mathcal P$, because $\mathbf Z^0$ better preserves the saliency-oriented one-positive behavior of the original CLIP model, whereas $\mathbf Z$ has been adapted through the cutting stage to include more contextual positives. Denote $\mathbf{Z}^0_i$ the confidence of label $i$ of all training data, the top-k samples of $\mathbf Z^0_i$ are selected as set $\mathcal P_i$, and
	\begin{equation}
		\mathcal P=\bigcup_{i=1}^C \mathcal P_i=\{p_1,p_2,p_3,\dots\},
	\end{equation}
	where $p_i$ is the serial number of the $p_i$th image in $\mathcal X'$. Note that $\mathcal P$ is a set and deduplicated. With salient images $\mathcal P$ and $\mathcal X'$, new training data $(\mathbf x_i^s,\mathbf z_i^s) \sim \mathcal X^s$ is constructed by
	\begin{align}
		\mathcal P'&=\{p'_1,p'_2,\dots\} \sim \mathcal P, \mathbf x_i \sim \mathcal X' \\
		\mathbf x_i^s&=\mathrm{Sew}(\mathbf x_i,\{\mathbf x_{p'_1},\mathbf x_{p'_2},\dots,\mathbf x_{p'_M}\};p,q) \\
		\mathbf z_i^s&=[z_{ij}^s]_{1 \leq j \leq C}, ~\text{where}~ z_{ij}^s= \max_{k \in \mathcal P'} z'_{kj},
	\end{align}
	where $z'_{kj}$ is the confidence of $j$th class of $k$th image. The function $\mathrm{Sew}(\mathbf x_i,\{\mathbf x_{p'_1},\mathbf x_{p'_2},\dots\};p,q)$ randomly selects multiple overlay images from $\{\mathbf x_{p'_1},\mathbf x_{p'_2},\dots,\mathbf x_{p'_M}\}$ with probability $p$, and sews the images on $\mathbf x$ while keeping their area ratio to $q$. After confidence blending, the label $\mathbf z^s_i$ of new $\mathbf x_i^s$ is a mixture of all $\mathbf z'$ of used images, thereby conforming to the multi-label confidence distribution for adaptation.
	%We freeze CLIP and train the adapter with constructed data by binary cross-entropy loss $\mathcal L_{\text{BCE}}(\mathbf p_i, \mathbf z'_i;\theta_{\{b,d,l\}})$.
	We freeze the backbone and train the adapter with constructed data by binary cross-entropy loss $\mathcal L_{\text{BCE}}$.
	
	\begin{figure}[t]
		\centering
		\includegraphics[width=\linewidth]{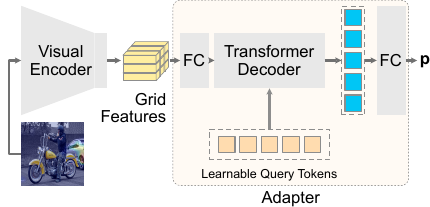}
		\caption{Illustrations of the model structures. The attention pooling layer of the VLM CLIP visual encoder is removed, and an adapter is attached for classification.}
		\label{fig:model}
	\end{figure}
	\subsection{Model Structures}\label{sect:model-structure}
	
	In \Cref{fig:model}, we show the simple structures of our frameworks. We first remove the attention pooling layer of the CLIP visual encoder (ResNet-101) to produce grid features $\mathbf F$ for the adapter to classify multi-label images. The adapter comprises parameter-sharing bottleneck $f_b$, vanilla Transformer decoder $f_d$, and a fully connected layer $f_l$ onto the grid features $\mathbf F$ produced by the visual encoder. The bottleneck transfers general features from the vision-language model to domain-specific features and the Transformer decoder uses learnable query tokens $\tau$ to decode the spatial domain-specific features to class-specific features for the fully connected layer $f_l$ to finally produce logits $\mathbf p$
	\begin{align}
		\mathbf F&=f_v(\mathbf x)=\begin{bmatrix}\mathbf f_{ij}\end{bmatrix}_{i=1,j=1}^{H,W}, \\
		\mathbf F'&=f_d(f_b(\mathbf F;\theta_b);\theta_d,\mathcal T_d)=[\mathbf f'_i]_{i=1}^C, \\
		\mathbf p&=f_l(\mathbf F';\theta_l),
	\end{align}
	where $\theta_{\{b,d,l\}}$ are learnable parameters, and $\mathcal T_d=\{t_i\}_{i=1}^C$ are learnable query tokens.
	
	During training, the visual encoder is frozen and only the adapter is optimized. The adapter is lightweight and has few parameters, which will be shown in \Cref{sect:ablation}.

	\subsection{Interactive Cutting and Sewing Optimization}
	
	To maintain the stability of pseudo labels and continuously eliminate the noise with the help of high-level knowledge of the model, we introduce the expectation-maximization algorithm.
	%from \citep{Cole2021singlePositive, chen2023semantic}.
	In the sewing stage, models, \ie, adapter $f_b,f_d,f_l$, and estimator $\mathbf Z$ in the framework are both optimized. In E-step, the estimator $\mathbf Z$ is optimized by
	\begin{align}
		\hat{\mathbf z}_i^s&=\arg\min_{\mathbf z}\xi(\hat{\mathcal F}(\mathbf x;\Theta),\mathbf z)
	\end{align}
	where $\Theta$ denotes learnable $\theta_{\{b,d,l\}}$ and $\mathcal T_d$. As $\mathbf z_i^s$ is a mixture of $\{\mathbf z'_j\}_{j \in \mathcal P'}$, multiple pseudo labels in $\mathcal P'$ will be optimized in one step
	\begin{equation}
		[z_{ij}^s]_{1 \leq j \leq C} \gets \hat{\mathbf z}_i^s, ~\text{where}~ z_{ij}^s= \max_{k \in \mathcal P'} z_{kj}.
	\end{equation}
	Instead of force updating $\mathbf Z$ with $\hat{\mathbf z}_i^s$, we implement it via back-propagating to maintain the stability.
	
	{Specifically, for each class $j$, the update of $z_{ij}^s$ is propagated to the source image that is most likely to contain class $j$ among all images involved in generating image $\mathbf x_i^s$. At the implementation level, this update is realized implicitly through back-propagation. We define the following loss for the E-step:
		\begin{equation}\label{eq:e-step-loss}
			\mathcal{L}_E(\mathbf{z}_i^s \mid \mathbf{p}_i; \Theta) = \mathcal{L}_{\mathrm{BCE}}\!\left(\mathbf{z}_i^s,\ \mathrm{Stop}(\mathbf{p}_i)\right),
		\end{equation}
		where $\mathrm{Stop}(\cdot)$ denotes the stop-gradient operator, $\mathbf p_i$ is the model prediction. Gradients are therefore back-propagated only through $\mathbf{z}_i^s$ and its upstream generation process. Through this mechanism, the update of each class is automatically routed to the most relevant source image via back-propagation.}
	
	In M-step, we optimize the model to predict the pseudo labels by loss $\mathcal L$
	\begin{equation}
		\mathcal L(\mathbf p_i \mid \mathbf z_i^s;\Theta)=\mathcal L_{\text{BCE}}(\mathbf p_i, \mathrm{Stop}(\mathbf z_i^s)),
	\end{equation} 
	By expectation-maximization, we interactively enhance the positive labels and suppress the sharp noise in $\mathbf Z$ to finally improve the models.
	
	\begin{algorithm}[htbp]
		\caption{Cut and Sew}\label{alg:sew}
		\KwData{$\mathcal X',p,q,L,M$}
		\KwResult{$\Theta$}
		\tcc{Process: Cut}
		$\mathbf Z^0=\begin{bmatrix} \mathbf z_i^0 \end{bmatrix}_{i=1}^N=\begin{bmatrix} \Phi(\mathbf x_i;\mathcal T) \end{bmatrix}_{i=1}^N$\;
		\For{$k=1$ \KwTo $M$}{
			$\hat{\mathcal X}'_k=\{\mathcal C(\mathbf x) \mid \mathbf x \in \mathcal X'\}$\;
			$\mathbf Z^k=\begin{bmatrix} \mathbf z_i^k \end{bmatrix}_{i=1}^N=\begin{bmatrix} \Phi(\mathbf x_i;\mathcal T) \end{bmatrix}_{\mathbf x_i \in \hat{\mathcal X}'_k}$\;
		}
		$\mathbf Z'=g(\mathbf Z)$\;
		\tcc{Process: Sew}
		\For {$i \gets 1$ \KwTo $C$}{
			$T_i =$ top-k images according to $\mathbf Z^0_{i}$\;
		}
		$T = \text{unique}\left(\bigcup_{i=1}^{C}T_i\right)$\;
		$q = \sqrt{\frac{q^2}{L}}$\;
		\tcp{Batch size = 1 for simplicity}
		\For{each batch image $\mathbf x \sim \mathcal X'$}{
			$\mathcal X^s \gets \{\}$\;
			\tcp{In practice, we also restrict the maximum number of stitched images}
			\While{$\mathrm{random}(0,1) \leq p \land |X| \leq L$}{
				$\mathcal X^s \gets \mathcal X^s \cup \mathbf x' \sim T$\;
				Resize $\mathbf x'$ by area $q \times~\text{size}(\mathbf x)$\;
				Paste $\mathbf x'$ into $\mathbf x$ at random position\;
			}
			$\mathbf z^s = \left[ z_{ij} \right]_{j=1}^C,~\text{where}~z_{ij}=\max_{i,\mathbf x_i \in \mathcal X^s} z_{ik}'$\;
			
			$\mathbf p = \hat{\mathcal F}(\mathbf x,\Theta)$\;
			\tcc{{M-step: optimize model parameters}}
			$\mathcal L_M = \mathcal L_{\text{BCE}}(\mathbf p, \mathrm{Stop}(\mathbf z^s))$\;
			\tcc{{E-step: update estimator $\mathbf Z$}}
			$\mathcal L_E = \mathcal L_{\text{BCE}}(\mathbf z^s, \mathrm{Stop}(\mathbf p))$\;
			Optimize $\Theta$ and $\mathbf Z$ with $\mathcal L_M+\mathcal L_E$;
		}
	\end{algorithm}

	We show the details of our framework's algorithm in \Cref{alg:sew}. It provides a clear overview of the whole training process including cutting and sewing operations. Denote $M$ as the number of sampling times. In practice, we restrict the maximum number of overlay images sewn to the main images by hyperparameter $L$.
	
	\begin{table*}[t]
		\small
		\centering
		\caption{mAP Comparison (\%) with state-of-the-art approaches on the VOC 2007/2012, MS-COCO, and NUS-WIDE datasets.}
		\begin{tabular}{c|c|c|cccc}
			\toprule
			\textbf{Supervision} & \textbf{Label} & \textbf{Method} & \textbf{VOC2012} & \textbf{VOC2007} & \textbf{MS-COCO} & \textbf{NUS-WIDE} \\
			\midrule
			\multirow{2}{*}{Fully Supervised} & \multirow{2}{*}{Full} & BCE-LS \citep{Cole2021singlePositive} & $91.6$ & $92.6$ & $79.4$ & $51.7$ \\
			& & BCE & $90.1$ & $91.3$ & $78.5$ & $50.7$ \\
			\midrule
			\midrule
			\multirow{10}{*}{Weakly Supervised} & \multirow{5}{*}{10\%} & Partial BCE \citep{Durand_2019_CVPR} & $81.3$ & $83.1$ & $63.2$ & $39.4$ \\
			&  & ASL \citep{Ridnik_2021_ICCV} & $-$ & $82.9$ & $69.7$ & $-$ \\
			&  & SST \citep{Chen2022SST} & $-$ & $81.5$ & $68.1$ & $-$ \\
			&  & SARB \citep{Pu2022SARB} & $-$ & $85.7$ & $71.2$ & $-$ \\
			&  & DualCoOp \citep{sun2022DualCoOp} & $-$ & $90.3$ & $78.7$ & $-$ \\
			\cmidrule{2-7}
			& \multirow{4}{*}{One} & ROLE \citep{Cole2021singlePositive} & $82.6$ & $84.6$ & $67.1$ & $43.2$ \\
			& & LL-R \citep{Kim2022largeLossML} & $89.7$ & $90.6$ & $72.6$ & $47.4$ \\
			& & $\mathrm{G}^2$NetPL \citep{abdelfattah2022g2netpl} & $89.5$ & $89.9$ & $72.5$ & $48.5$ \\
			& & Scob \citep{chen2023semantic} & $88.5$ & $89.7$ & $74.8$ & $-$ \\
			\midrule
			\multirow{6}{*}{Unsupervised}& \multirow{6}{*}{None}  & DualCoOp \citep{sun2022DualCoOp} & $-$ & $-$ & $-$ & $43.6$ \\
			&  & CDUL \citep{Abdelfattah2023CDUL} & $88.6$ & $89.0$ & $69.2$ & $44.0$ \\
			&  & TagCLIP \citep{lin2024tagclip} & $-$ & $92.8$ & $68.8$ & $-$ \\
			&  & CCD \citep{kim2025classifier} & $90.1$ & $91.0$ & $70.3$ & $44.5$ \\
			&  & \cellcolor{lightgray} \textbf{Ours (ResNet-101)} & \cellcolor{lightgray} \textbf{$91.4$} & \cellcolor{lightgray} \textbf{$91.8$} & \cellcolor{lightgray} \textbf{$72.5$} & \cellcolor{lightgray} \textbf{$45.6$} \\
			&  & \cellcolor{lightgray} {\textbf{Ours (ViT-L/14)}} & \cellcolor{lightgray} {\textbf{$90.0$}} & \cellcolor{lightgray} {\textbf{$92.8$}} & \cellcolor{lightgray} {\textbf{$79.8$}} & \cellcolor{lightgray} {\textbf{$44.6$}} \\
			\bottomrule
		\end{tabular}%
		\label{tab:mAP-metrics}%
	\end{table*}
	
	\begin{table}[t]
		\small
		\centering
		\caption{mAP of different label correction strategies on the MS-COCO dataset.}
		\begin{tabular}{c|cc}
			\toprule
			\textbf{Methods} & \textbf{Training set} & \textbf{Testing set}  \\
			\midrule
			No correction & $67.5\phantom{(+0.0)}$ & $71.7\phantom{(+0.0)}$  \\
			Hard correction & $64.1(-3.4)$ & $71.7(+0.0)$ \\
			\rowcolor{lightgray} Ours & $\mathbf{67.5(-0.0)}$ & $\mathbf{72.5(+0.8)}$ \\
			\bottomrule
		\end{tabular}%
		\label{tab:correction}%
	\end{table}%
	
	\begin{table}[t]
		\small
		\centering
		\caption{Effect of different components. *: mAP on training data is evaluated in the cutting stage.}
		\begin{tabular}{c|cc|cc}
			\toprule
			\textbf{Estimate} & \textbf{Correct} & \textbf{Adapt} & \textbf{mAP} & $\Delta$ \\
			\midrule
			&  &  & $85.1$* & $-$ \\
			$\checkmark$ &  &  & $88.3$* & $\uparrow 3.2$ \\
			\midrule
			$\checkmark$ & $\checkmark$  &  & $88.6$ & $-$ \\
			$\checkmark$ &  & $\checkmark$  & $91.1$ & $\uparrow 2.5$ \\
			\rowcolor{lightgray} $\checkmark$ & $\checkmark$  & $\checkmark$  & $91.4$ & $\uparrow 2.8$ \\
			\bottomrule
		\end{tabular}%
		\label{tab:ablation-component}%
	\end{table}%

	\section{Experiments}\label{sect:expsum}
	
	\subsection{Experiment Setup}
	
	\subsubsection{Datasets}
	Following previous works \citep{Abdelfattah2023CDUL,sun2022DualCoOp,chen2023semantic}, we conduct experiments on four widely-used benchmarks: PASCAL VOC 2007 \citep{pascal-voc-2007}, PASCAL VOC 2012 \citep{pascal-voc-2012}, Microsoft COCO 2014 \citep{Lin2014MSCOCO}, and NUS-WIDE \citep{Chua2009nus-wide}. These datasets are fully labeled for multi-label recognition. PASCAL VOC contains 20 categories, with 1.4 labels per image on average. Microsoft COCO contains 80 categories and is challenging due to complex scenes with multiple objects, each labeled with 2.9 categories on average. NUS-WIDE contains 81 categories, about 150K training images and 60K evaluation images.
	
	For our settings, all labels of the training set are dropped before training, which means the framework has only access to images $\mathcal X'$. For the methods trained on fully labeled data, we keep the original labels for training. For the methods that require partial labels, the labels are randomly dropped, leaving $10\%$ labels to satisfy the requirement following previous works \citep{Durand_2019_CVPR, Abdelfattah2023CDUL}. For the methods trained on single labels, we randomly select one label for each image following previous works \citep{Cole2021singlePositive,chen2023semantic}. The official validation set is used for testing.
	
	\subsubsection{Implementation and Evaluations}
	In cutting stage, following \citep{Abdelfattah2023CDUL}, we use $\text{ResNet-50}\times\text{64}$ CLIP for only zero-shot learning in first stage. Images are sampled $20$ times for the estimator to generate $\mathbf Z^\cdot$. $\alpha$ is set to $0.1$. In the sewing stage, we freeze ResNet-101 CLIP and train the adapter for $1$ epoch with AdamW \citep{adamw} optimizer. The learning rates of the adapter and estimator are set as $0.001$, and $0.01$ respectively with a weight decay of $0.01$. The batch size is set as $16$. $k$ is set as $200$. The hyperparameter $\beta$ is set as $1$ for the VOC dataset, and $2$ for the MS-COCO dataset, NUS-WIDE dataset. {$p$ and $q$ are both set as $0.6$.}
	{We additionally provide experiments using a ViT-based CLIP backbone (ViT-L/14@336px). However, for fair comparison with mainstream methods, ResNet-101 is used as the default backbone in comparative evaluations.}
	Following previous works~\citep{Abdelfattah2023CDUL,sun2022DualCoOp,chen2023semantic}, we set the resolution of images as $448 \times 448$ with data augmentations for fair comparisons and adopt the mean Average Precision (mAP) as the evaluation metrics following these works.

	\subsection{Comparison with State-of-The-Art Approaches} \label{sect:exp}
	
	We compare our framework on the VOC 2007, VOC 2012, MS-COCO, and NUS-WIDE datasets with 13 state-of-the-art methods: 1) 2 fully supervised methods BCE-LS \citep{Cole2021singlePositive} and BCE for reference. 2) 9 weakly supervised methods, including Partial BCE \citep{Durand_2019_CVPR}, ASL \citep{Ridnik_2021_ICCV}, SST \citep{Chen2022SST}, SARB \citep{Pu2022SARB}, DualCoOp \citep{sun2022DualCoOp}, ROLE \citep{Cole2021singlePositive}, LL-R \citep{Kim2022largeLossML}, $\text{G}^2\text{NetPL}$ \citep{abdelfattah2022g2netpl}, and Scob \citep{chen2023semantic}. 3) 4 unsupervised methods DualCoOp \citep{sun2022DualCoOp}, CDUL \citep{Abdelfattah2023CDUL}, TagCLIP \citep{lin2024tagclip} and CCD \citep{kim2025classifier}. Following \citep{Abdelfattah2023CDUL,chen2023semantic}, we use $10\%$ labels for those trained with partial labels. The results are reported in \Cref{tab:mAP-metrics}. Compared with unsupervised methods, our proposed method significantly outperforms others on 4 datasets by $2.8\%$, $2.8\%$, $3.3\%$, and $1.6\%$ respectively, confirming the effectiveness of our framework. {In addition, we report results with a ViT-based CLIP backbone in \Cref{tab:mAP-metrics}. The ViT variant achieves the best performance on MS-COCO and VOC 2007, demonstrating generalization across backbones and stronger global representations. On VOC 2012 and NUS-WIDE, the ResNet-based model remains competitive, indicating that the proposed response estimation and fusion strategy is not backbone-dependent. These results suggest that the performance gains mainly come from our method rather than the choice of visual encoder.} Moreover, our framework even \textbf{surpasses all weakly supervised methods} on the VOC 2012 and VOC 2007 datasets, and most of the 13 methods on the MS-COCO and NUS-WIDE datasets. This proves that the quality of pseudo labels generated by ours is comparable to the ground truth annotations. Considering that these methods require partial labels to satisfy random sampling, which is hard to achieve in practice, we believe that our framework exceeds them in both performance and feasibility. Finally, our framework narrows the performance gap with fully supervised methods and successfully eliminates the labor consumption of labeling data.

	\subsection{Ablation Studies}\label{sect:ablation}
	
	\begin{table}[t]
		\small
		\centering
		\caption{Effect of estimated $\mathbf Z$. The mAP is continuously improving during training. C.: Cutting. S.: Sewing}
		\begin{tabular}{c|ccc|c}
			\toprule
			\textbf{Datasets} & \textbf{Init.} & \textbf{C. Stage} & \textbf{S. Stage} & $\Delta$ \\
			\midrule
			VOC2012 & $85.1$ & $88.3$ & $89.4$ & $\uparrow 4.3$ \\
			VOC2007 & $84.9$ & $88.6$ & $90.4$ & $\uparrow 5.5$ \\
			MS-COCO & $64.9$ & $70.6$ & $70.7$ & $\uparrow 5.8$ \\
			% NUS-WIDE & $39.0$ & $39.0$ & $39.0$ & $\uparrow 0.0$ \\
			\bottomrule
		\end{tabular}%
		\label{tab:estimating}%
	\end{table}%

	\begin{table}[t]
		\small
		\centering
		\caption{Effect of sampling iterations $M$ on VOC2007. The performance improves with more diverse local views and saturates at $M=20$.}
		\begin{tabular}{c|cccccc}
			\toprule
			$M$ & 1 & 5 & 10 & 20 & 30 & 50 \\
			\midrule
			ResNet-101 & 89.5 & 90.7 & 92.0 & \textbf{92.5} & 92.6 & 92.7 \\
			{ ViT-L/14} & { 87.4} & { 89.7} & { 90.7} & { \textbf{92.2}} & {91.2} & {89.9} \\
			\bottomrule
		\end{tabular}
		\label{tab:ablation-N}
	\end{table}
	
	\begin{table}[t]
		\small
		\centering
		\caption{The numbers of inversions with different label correction strategies on the training set.}
		\begin{tabular}{c|c|cc}
			\toprule
			\textbf{Datasets} & \textbf{Original} & \textbf{Hard Correction} & \textbf{Ours} \\
			\midrule
			VOC 2012 & $80.38\text{K}$ & $127.5\text{K}~(\uparrow 58\%)$ & $80.33\text{K}~(\downarrow)$ \\
			VOC 2007 & $13.00\text{K}$ & $19.73\text{K}~(\uparrow 51\%)$ & $12.96\text{K}~(\downarrow)$ \\
			MS-COCO & $17.61\text{M}$ & $47.28\text{M}~(\uparrow 168\%)$ & $17.60\text{M}~(\downarrow)$ \\
			% NUS-WIDE & 0000 & 0000 & 0000 (+0) \\
			\bottomrule
		\end{tabular}%
		\label{tab:inversion}%
	\end{table}%
	
	\begin{table}[t]
		\small
		\centering
		\caption{Ablation study on hyperparameter $\beta$ on VOC2007.}
		\begin{tabular}{c|ccccc}
			\toprule
			$\beta$ & 1 & 2 & 3 & 4 & 5 \\
			\midrule
			mAP (\%) & 91.8 & 91.3 & 91.1 & 91.3 & 91.2 \\
			\bottomrule
		\end{tabular}
		\label{tab:beta}
	\end{table}

	\subsubsection{Effect of Components} 
	To study the effect of proposed components in two stages, we respectively ablate them and measure mAP performance on the VOC 2012 dataset, which is reported in \Cref{tab:ablation-component}. In the upper part of \Cref{tab:ablation-component}, the quality of confidence (pseudo labels) significantly decreases without the estimator, further corroborating our observations in \Cref{sect:method}, that CLIP is not credible in multi-label tasks. The estimator mitigates the limitations of CLIP and generates high-quality pseudo labels. In the lower part of the table, we show that trivially fine-tuning a CLIP w/o our adaptation reaches $88.6$, which is almost identical to cutting stage performance $88.3$. The reason is that CLIP still follows the ordinary one-positive distributions. Naively distilling CLIP preserves its single-positive bias, limiting multi-label performance. Meantime, the correction enhances model accuracy, and we will discuss it in the next part.
	
	\subsubsection{Effect of Response Estimating} 
	We report the quality of the pseudo labels at different stages of training in \Cref{tab:estimating}. After the cutting stage, the quality of pseudo labels improves noticeably, clearly confirming the effectiveness of the estimator. In the sewing stage, the quality continues to rise consistently with interactive learning, taking full advantage of the model's knowledge.

	\textbf{Effect of $M$, $\alpha$ and {$\beta$}.} We further investigate the sensitivity of performance to the number of sampling iterations $M$ and $\alpha$ in \Cref{eq:7}. As shown in \Cref{tab:ablation-N}, mAP consistently improves as $M$ increases. A larger $M$ introduces more random crops, but potential noise from oversampling is effectively controlled by the suppression strategy, ensuring stable response estimation. Performance nearly converges at $M=20$, and increasing $M$ further yields only marginal gains while incurring higher computational cost. {A similar trend is also observed with the ViT backbone: the performance improves as $M$ increases and reaches the best result at $M=20$, which indicates that the cutting strategy is also effective for ViT-based architectures.} In \Cref{fig:ablation-pq}, we study $\alpha$ in \Cref{eq:7}. When $\alpha$ is small, the suppression of noise in the logits is insufficient, while when $\alpha$ is too large, overly strong suppression also weakens discriminative information, leading to performance degradation. As a result, $\alpha = 0.1$ achieves the best trade-off. {We further study the sensitivity to $\beta$ in \Cref{eq:g}, with results reported in \Cref{tab:beta}. The performance differences across different $\beta$ values are relatively small. The best result is achieved at $\beta=1$, which is used as the default setting. As $\beta$ increases, mAP shows a slight decrease, suggesting that correcting only the most confident category is already sufficient on VOC2007, while larger $\beta$ may introduce less reliable categories. Overall, the results indicate that the proposed correction is not highly sensitive to $\beta$.}
		
		It should be emphasized that even if there is a significant improvement in the quality of pseudo labels, according to \Cref{tab:ablation-component}, such improvement still does not benefit the multi-label recognition of CLIP in the absence of the multi-object blend adaptation. This indicates that the quality improvement is still subject to the underlying distribution of vanilla CLIP output, and that the sewing stage plays a crucial role in changing the properties of CLIP.

		\subsubsection{Effect of Multi-object Blending}\label{sect:ablation-pq}
\begin{figure}[t]
			\centering
			\includegraphics[width=.98\linewidth]{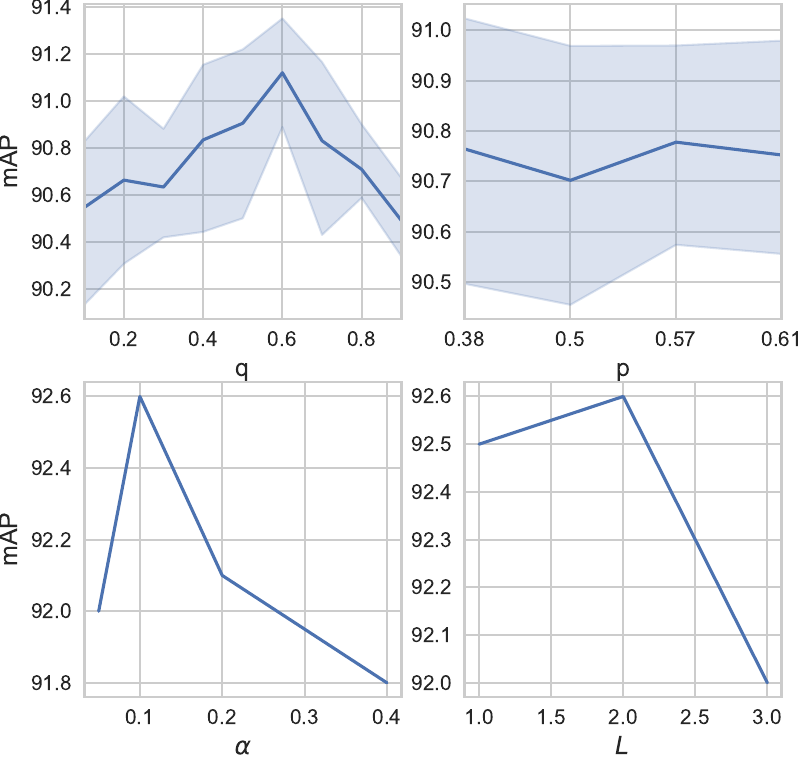}
			\caption{Ablation studies to hyperparameter $p$, $q$, $\alpha$ and $L$.}
			\label{fig:ablation-pq}
		\end{figure}

		{We ablate hyperparameters $p$, $q$, and $L$ in multi-object blend adaptation, where $p$ is the probability to determine the number of images to be sewed, $q$ is the ratio of areas of all sewed images to the background images, and $L$ is the maximum number of stitched images.} The results are reported in \Cref{fig:ablation-pq}.
		
		{For the probability $p$ and number $L$, the performance decreases gradually as the probability increases, \ie, the number of images pasted in images increases. Too many images may cover the original objects, corrupt the object features, and affect the training. As in the ablation experiments for image ratio $q$, the performance shows a trend of increasing and then decreasing as the $q$ increases, which is intuitive. A too-large object may cover the original objects in images, while a too-small object is no longer salient for adaptation, both of which can affect the recognition. Although differences are observed, the performance is not sensitive to the hyperparameters, indicating that the proposed method is robust in most scenarios. In practice, sewing 2 or 3 images in total is sufficient to adapt CLIP for multi-label recognition.}
		
		\subsubsection{Label Correction Strategies} \label{sect:label-correction-ablation}
		We compare our label correction with the hard correction strategy similar to \citep{sohn2020fixmatch,Zhang2021Flexmatch}, which sets the confident categories meeting thresholds as positive labels $1$, leaving the rest unchanged. The results are reported in \Cref{tab:correction}.
		%When labels are not adjusted, the training behavior approximates distillation.
		
		With hard correction, the performance of the model does not improve on the test set and decreases on the training set.
		Although such methods have been verified to be effective in semi-supervised learning, they are not suitable for multi-label learning. The degradation of the performance on the training set indicates they introduce noise in the data. As our correction strategy keeps the relative order of the confidence, the performance on the training set is not affected, however, it changes the confidence distribution of the pseudo labels, encouraging the correct labels in model learning, thus significantly improving the model performance on the testing set.
		
		We additionally count the number of inversions in the pseudo labels before and after the correction in \Cref{tab:inversion}. In detail, the number of inversions {increases} when the confidence (pseudo labels) of a negative sample is higher than that of a positive, which can impair the model's learning of features. The results show that the number of inversions increases significantly with the hard correction, while our correction strategy keeps the number of inversions almost unchanged.
		
		{\textbf{Effect of Correction Function $g(\cdot)$.}
			We further conduct ablation studies to justify the choice of the square-root rescaling function in \Cref{eq:g}.
			Specifically, we compare the following alternative correction functions:
			\begin{equation}\label{eq:g-hard}
				\small
				g_{\mathrm{hard}}(z_{ij}) =
				\begin{cases}
					\mathcal S^{-1}(0.95), & j = \arg\max_k z_{ik}, \\
					\mathcal S^{-1}(0.05), & j = \arg\min_k z_{ik}, \\
					z_{ij}, & \text{otherwise}.
				\end{cases}
			\end{equation}
			\begin{equation}\label{eq:g-lin}
				\small
				g_{\mathrm{linear}}(z_{ij}) =
				\begin{cases}
					\mathcal S^{-1}(\alpha + (1-\alpha)\,\mathcal S(z_{ij})), & z_{ij}~\text{in top-}\beta~\text{of}~\mathbf z_i, \\
					z_{ij}, & \text{otherwise}.
				\end{cases}
			\end{equation}
			The square-root function is monotonic but concave, exhibiting sub-linear growth, which reduces the relative gap between low- and high-confidence predictions.
			This property allows low-confidence but informative predictions to have a more noticeable impact during optimization.
			We compare different functions on VOC2007, as shown in \Cref{tab:ablation-scaling}.
			From the results, we observe that the square-root function consistently outperforms the alternatives.}

		\begin{table}[t]
			\small
			\centering
			\caption{Effect of different correction functions on VOC2007. Square-root function achieves the best performance.}
			\begin{tabular}{c|cccc}
				\toprule
				$g(\cdot)$ & hard & soft (Ours) & linear & no correction \\
				\midrule
				mAP (\%) & 91.2 & \textbf{92.6} & 92.2 & 92.2 \\
				\bottomrule
			\end{tabular}
			\label{tab:ablation-scaling}
		\end{table}
		
		\begin{figure}[t]
			\centering
			\includegraphics[width=\linewidth]{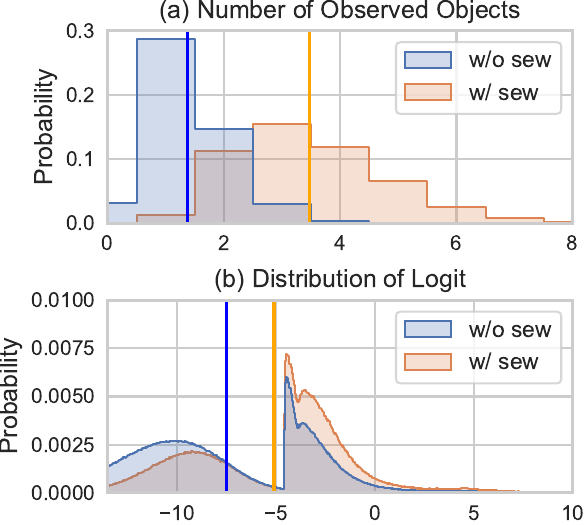}
			\caption{The distributions of observed objects and logits w/ and w/o sew. (a) The number of observed objects per image by CLIP on the MS-COCO datasets, which has $2.9$ labels per image on average. After sewing, the number of observed objects manages to get out of the limit and reach the average of the real data.
				(b) The numerical distribution of logits. Our method does not change the distribution of values, ensuring fast adaptation. The {\color{blue}blue} and {\color{orange}orange} line marks the mean values {\color{blue}w/o} and {\color{orange}w/} sewing respectively.}
			\label{fig:sew-distribution}
		\end{figure}
		
		\subsection{Discussions on Multi-label CLIP}
		\subsubsection{Limited Responses of CLIP and Effect of Blending}
		We analyze the distributions of prediction w/ and w/o sew to show its real effects on the MS-COCO dataset. From \Cref{fig:sew-distribution}(a) we confirm again that CLIP responds to only one object, \ie, the salient one, in most cases, as the average object observed by CLIP is almost $1$.
		As the COCO dataset is a multi-label dataset, predicting only one object leads to a huge amount of false negatives, and the single-positive responses suppress other objects which introduces more noise. This explains why directly distilling CLIP fails to yield improvements on multi-label tasks. With our framework, the average number of observed objects per image is significantly increased to $3.4$, which is closer to the real distribution of the MS-COCO dataset, \ie, $2.9$, indicating our framework overcomes CLIP's limited multi-object response. While the number of observed objects changes a lot, the numerical size of distilled logits is stable to maintain the parameter distribution of CLIP for fast knowledge adaptation, which describes the short adaptation time of our framework, \ie, one epoch.

		% Should we add more visualization from the supplementary material? Yes, I added. - CC
		\begin{figure}[t]
			\centering
			\includegraphics[width=\linewidth]{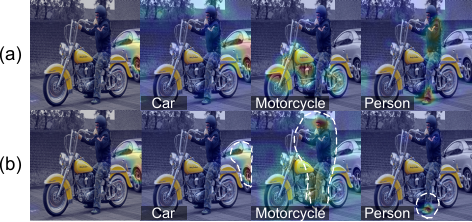}
			\caption{The class activation maps by (a) ours and (b) CLIP on a complex image, containing ``person'', ``motorcycle'', and ``car''. The overlay color indicates the areas of focus for recognizing the categories. We mark the inconspicuous and wrong activation areas with dotted lines in the figure.}
			\label{fig:cam}
		\end{figure}

		\begin{table}[t]
			\small
			\centering
			\caption{The top 10 easiest and 10 hardest categories to recognize for CLIP on the MS-COCO dataset.}
			\begin{tabular}{cc|cc}
				\toprule
				\textbf{No.} & \textbf{Category} & \textbf{No.} & \textbf{Category} \\
				\midrule
				1 & person & 80 & mouse \\
				2 & airplane & 79 & toaster \\
				3 & bus & 78 & snowboard \\
				4 & baseball bat & 77 & baseball glove \\
				5 & giraffe & 76 & spoon \\
				6 & train & 75 & potted plant \\
				7 & cake & 74 & hair drier \\
				8 & elephant & 73 & refrigerator \\
				9 & clock & 72 & handbag \\
				10 & boat & 71 & toothbrush \\
				\bottomrule
			\end{tabular}%
			\label{tab:bias}%
		\end{table}%

		\begin{figure}[t]
			\centering
			\includegraphics[width=0.96\linewidth]{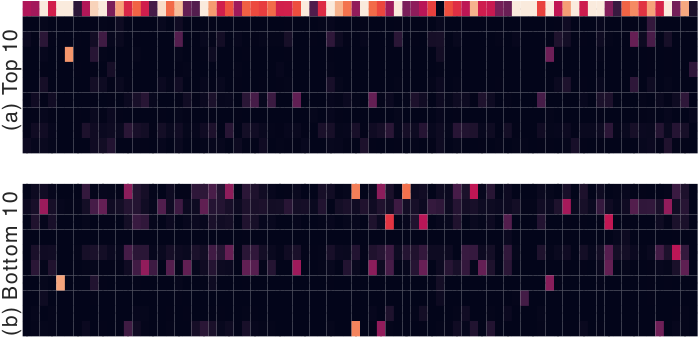} % to align with figure on the left
			\caption{The co-occurrence probabilities between top/bottom 10 categories (listed in \Cref{tab:bias}) and others. A brighter color indicates that the corresponding top/bottom categories are more likely to appear with that category, while a color not bright enough indicates an unstable relationship.}
			\label{fig:co-relation}
		\end{figure}
		
		\begin{figure*}[t]
			\centering
			\includegraphics[width=0.9\linewidth]{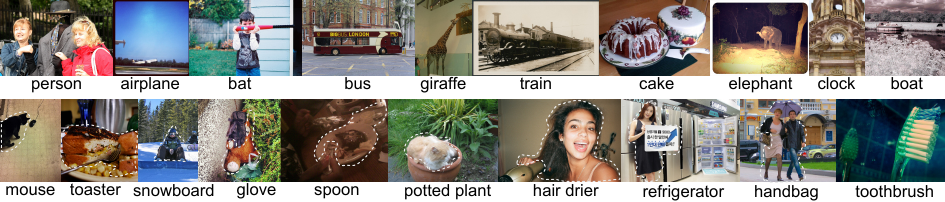}
			\caption{The top-1 confident images of top/bottom 10 categories on the MS-COCO dataset. We mark the main objects with dotted lines in the bottom 10. Despite our selection of CLIP's most confident images, these categories remain unremarkable, indicating the inherent bias.}
			\label{fig:top-image}
		\end{figure*}

		\subsubsection{{Why Does CLIP Struggle to Recognize Multiple Objects?}} \label{sect:cam}
\begin{figure*}[t]
			\centering
			\includegraphics[width=0.9\linewidth]{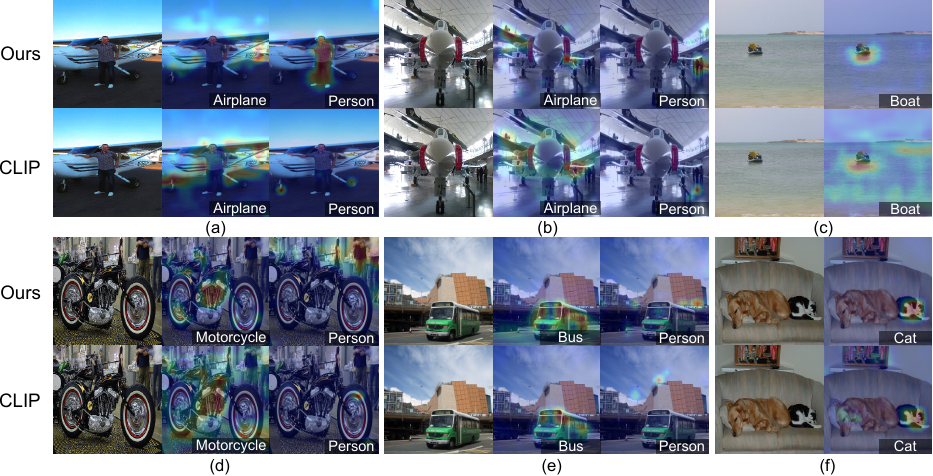}
			\caption{The class activation maps by ours and vanilla CLIP on various images. Our method successfully finds the targets and the CAM is more accurate. The CLIP tends to focus on the dominating objects while encoding the rest as the context of them. a), b), d), e) CLIP encodes ``person'' as the context of the main objects, and ignores discovered context when recognizing ``person''. c) The whole image is associated with ``boat''. f) The irrelevant dog is counter-intuitively activated when recognizing ``cat''.}
			\label{fig:more-cam}
		\end{figure*}

		We visualize the class activation maps produced by ours and CLIP in \Cref{fig:cam}. From the figure, we further observe the misbehavior of CLIP and the advantages of our framework. In the CAM of ``car'', the activation of our framework on the target is much higher than that of CLIP, confirming the superiority of our framework. Although CLIP responds extremely to the ``motorcycle'', it incorrectly recognizes the person as a part of it, while our framework accurately focuses on the target only with high activation. When recognizing ``person'', CLIP instead responds too little and barely finds the person from the image. To further illustrate the phenomenon, we provide more visualizations in \Cref{fig:more-cam}. From the figure, we find that
		\begin{itemize}
			\item When there are salient objects (plane, motorcycle, bus) dominating the images, the CLIP's focus on other objects, \ie, ``person'', simply avoids areas activated by them and locates at strange random positions, though it already finds them out as context of the dominating objects.
			\item In \Cref{fig:more-cam}(c), the CLIP classifies the ``boat'' while noticing its surroundings, which indicates that CLIP looks at the image as a whole.
			\item In \Cref{fig:more-cam}(f), when two key objects exist, the CLIP classifies ``cat'' while also focusing on the ``dog'', indicating that it hardly encodes objects separately.
		\end{itemize}

		We believe this phenomenon reflects, in part, the internal mechanism of CLIP. It further validates the observation in \Cref{fig:clip-zero-shot-behavior} that CLIP recognizes images starting with salient objects and encodes their associated contexts together as visual features. The decoupled design of encoders and image-caption data obtained from the web enables the CLIP visual encoder to focus on the salient object and its related context when encoding the whole image for similarity comparison with textual features. This is a good strategy for ``image-caption'' pre-training, but leads to an imbalance detrimental to multi-label learning. Our framework successfully mitigates this and enhances the multi-label recognition performance of CLIP.

		\subsubsection{Exploring the bias of CLIP}\label{sect:bias}
		% This may be merged with the previous part. No, I think it's better to separate them. - CC
		To further uncover the bias of CLIP recognition, we list the easiest and hardest 10 categories for CLIP to recognize in \Cref{tab:bias}, ordered by top-1 accuracy. We observe that the top 10 categories are all salient or not significantly related to other categories, while the bottom 10 are inconspicuous or concomitant with other categories.
		
		To illustrate that, we show the co-occurrence probabilities in \Cref{fig:co-relation} and the top-1 confident images of each category in \Cref{fig:top-image}.
		
		\begin{itemize}
			\item In \Cref{fig:co-relation}(a), the top 10 categories have little relevance to other categories except for the top 1 (``person''), which has strong relevance to many categories to make it salient.
			\item In \Cref{fig:co-relation}(b), the bottom 10 categories have unstable weak associations with a large number of other categories, which means they are usually secondary objects, leading to CLIP concentrating on other salient objects.
			\item \Cref{fig:top-image} provides further evidence to show the effects of the salient objects on other inconspicuous objects existing in the image. CLIP focuses on those salient and treats the ground-truth targets as concomitant.
		\end{itemize}
		
		We can also clearly observe from \Cref{fig:top-image} that this bias makes CLIP tend to focus on the salient objects, which effectively guarantees the correctness of our multi-object blend adaptation.

		\subsubsection{Visualization of Features}
\begin{figure}[t]
			\centering
			\includegraphics[width=\linewidth]{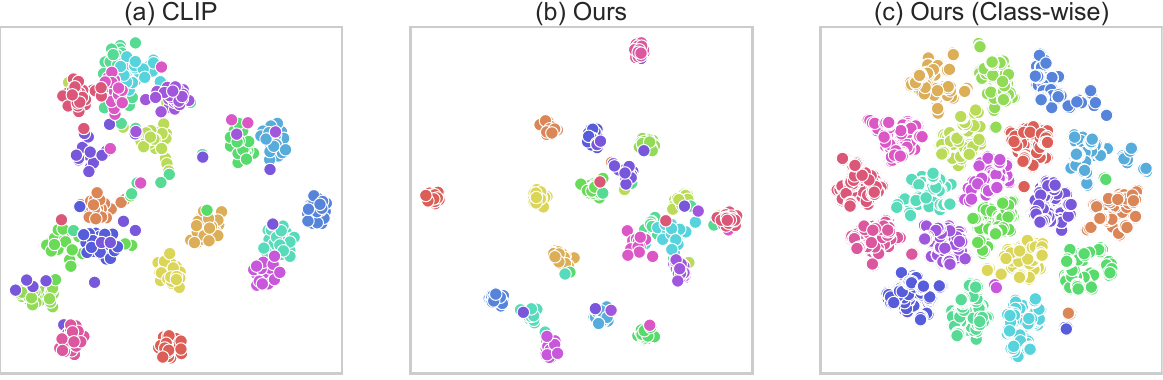}
			\caption{The visualization of features by CLIP and ours via t-SNE. (a) Overlap is observed in 600 image features by CLIP. (b) Our method produces more significant features for 600 images. (c) The class-wise features of 600 images $\times$ 20 classes by our framework, further distinguish different categories.}
			\label{fig:visual-features}
		\end{figure}

		We visualize the features of $600$ images by CLIP and ours in \Cref{fig:visual-features}. Serious overlap can be observed in \Cref{fig:visual-features}(a), which indicates that the visual encoder of vanilla CLIP cannot distinguish several categories when the images contain multiple objects, leading to obscurity in classification. The features produced by ours are more concentrated and significant. Our class-wise features in \Cref{fig:visual-features}(c) further overcome the confusion, which are less overlap and more distinguishable.
		
		\begin{table}[t]
			\small
			\centering
			\caption{Efficiency of parameters and inference time.}
			\begin{tabular}{c|cc}
				\toprule
				\textbf{Method} & \textbf{\# of Params} & \textbf{Inference Time} \\
				\midrule
				CLIP & $120.0\text{M}$ & $55.8\text{ms}$ \\
				ResNet-101 & $44.5\text{M}$ & $11.4\text{ms}$ \\
				\rowcolor{lightgray} Ours & $44.9\text{M}$ & $14.6\text{ms}$ \\
				\bottomrule
			\end{tabular}%
			\label{tab:efficiency}%
		\end{table}%
		\subsubsection{Efficiency}\label{sect:eff}
		To analyze the efficiency of our frameworks, we report the inference time and the number of total parameters in \Cref{tab:efficiency}, including ours, vanilla CLIP \citep{Radford2021CLIP} and ResNet-101 \citep{He2016resnet} as baseline. For a fair comparison, we resize the input to resolution $448 \times 448$ and run all methods on the MS-COCO dataset \citep{Lin2014MSCOCO} to count the average time. The experiments are conducted on a single NVIDIA 3090 GPU. From the table, we can find that the parameters and inference time of our framework are almost identical to the widely-used ResNet-101 ($+0.4\text{M}$, $+3.2\text{ms}$), much lower and faster than a vanilla CLIP with backbone ResNet-101. This high efficiency ensures that our framework can be fluently migrated to existing methods and systems to free them from human annotating.

		\section{Conclusions and Limitations} \label{sect:conclusion}
		
		In this paper, we analyze the behavior of vision-language models (VLMs) on multi-label tasks to discuss their one-positive limitation. From the experiments, we observe that these models often respond primarily to the most iconic object while omitting other contextual positive objects, limiting their performance on multi-label tasks where objects are relatively more independent. Based on this observation, we propose an unsupervised framework that adapts VLMs from iconic recognition toward inclusive understanding, enabling label-free multi-label image recognition. Our approach consists of two key stages, ``cutting'' and ``sewing'', which accurately adjust the distribution of model predictions to adapt vision-language models for multi-label recognition without labels. Extensive experiments show that our framework significantly outperforms existing unsupervised approaches on four public datasets, even surpassing several representative weakly supervised baselines.
		
		While our proposed approach extracts meaningful clues from the pre-trained vision-language models, the upper bound of this unsupervised learning scheme is still restricted by the latent knowledge in these models. In future work, we would like to explore this learning scheme from larger multi-modal contrastive models or large-scale language models.
		
		\section*{Statements and Declarations}
		
		\textbf{Funding.}
		This work is partially supported by grants from the National Natural Science Foundation of China (No.62132002), Guizhou Provincial Major Scientific and Technological Program (Qiankehe Zhongda [2025] No. 032), Beijing Nova Program (No.20250484786), and the Fundamental Research Funds for the Central Universities.
		
		\textbf{Competing Interests.}
		The authors have no relevant financial or non-financial interests to disclose.
		
		\textbf{Ethics Approval.}
		This article does not contain any studies with human participants or animals performed by any of the authors.
		
		\textbf{Consent to Participate.}
		This study does not involve experiments requiring informed consent from participants; therefore, this item is not applicable.
		
		\textbf{Consent for Publication.}
		All authors have approved the final manuscript and consent to its publication.
		
		\textbf{Author Contributions.}
		Cheng Chen and Jingyu Zhou contributed equally to this work. Cheng Chen contributed to conceptualization, methodology, experiments, and manuscript writing. Jingyu Zhou contributed to methodology, experiments, validation, and visualization. Yifan Zhao contributed to conceptualization, supervision, funding acquisition, and manuscript revision. Jia Li contributed to supervision, funding acquisition, resources, and manuscript revision. All authors read and approved the final manuscript.
		
		\textbf{Data Availability.}
		The datasets used or analyzed in the current study are available from the original sources:
		PASCAL VOC 2007/2012~\citep{pascal-voc-2012},
		Microsoft COCO 2014~\citep{Lin2014MSCOCO}, and
		NUS-WIDE~\citep{Chua2009nus-wide}.
		The corresponding dataset pages are
		\href{http://host.robots.ox.ac.uk/pascal/VOC/}{PASCAL VOC},
		\href{https://cocodataset.org/}{COCO}, and
		\href{https://huggingface.co/datasets/Lxyhaha/NUS-WIDE}{NUS-WIDE}.

		% \section*{Appendix}
		% appendix


		\bibliographystyle{spbasic}      % basic style, author-year citations
		\bibliography{refs}   % name your BibTeX data base

		\balance
	\end{document}